\documentclass[letterpaper, 10 pt, conference]{ieeeconf}  % Comment this line out if you need a4paper
\IEEEoverridecommandlockouts                              % This command is only needed if 
\usepackage{amsmath} % assumes amsmath package installed
\usepackage{amssymb}  % assumes amsmath package installed
\usepackage{siunitx}
\usepackage{pifont}
\newcommand{\xmark}{\ding{55}}%
\usepackage{comment}
\usepackage{booktabs}
\usepackage{tikz}
\usepackage{acronym}
\usepackage[hidelinks]{hyperref}
\usepackage{float}% If comment this, figure moves to Page 2
\usepackage{lipsum}

\usepackage{dsfont}

\usepackage{balance} % Used to 'balance' the references at the end of the paper so it looks tidy
\title{\LARGE \bf
MEM: Multi-Modal Elevation Mapping for Robotics and Learning
}
\author{Gian Erni$^\dagger$, Jonas Frey$^\dagger$, Takahiro Miki$^\dagger$, Matias Mattamala$^\ddagger$, Marco Hutter$^\dagger$%
}

\usepackage[abbreviations]{glossaries-extra}

\glssetcategoryattribute{abbreviation}{nohyperfirst}{true}

\newabbreviation{ros}{ROS}{Robot Operating System}
\newabbreviation{fov}{FoV}{Field of View}
\newabbreviation{gpu}{GPU}{Graphics Processing Unit}
\newabbreviation{cpu}{CPU}{Central Processing Unit}
\newabbreviation{cuda}{CUDA}{Compute Unified Device Architecture}
\newabbreviation{mem}{MEM}{Multi-Modal Elevation Map}
\newabbreviation{erf}{ERF Net}{Efficient Residual Factorized ConvNet}
\newabbreviation{cnn}{CNN} {Convolutional Neural Network}

\def\secref#1{Sec.~\ref{#1}}

\def\figref#1{Fig.~\ref{#1}}
\def\tabref#1{Tab.~\ref{#1}}
\def\eqref#1{Eq.~(\ref{#1})}

\newcommand{\etal}{et al.}

\begin{document}

\twocolumn[{%
\renewcommand\twocolumn[1][]{#1}
\maketitle
\thispagestyle{empty}
\pagestyle{empty}
\begin{center}
    \centering
    % left bottom right up
	\begin{minipage}{\textwidth}
        \begin{figure}[H]
        \centering
            \begin{minipage}{\textwidth}
                \includegraphics[width=1.0\textwidth]{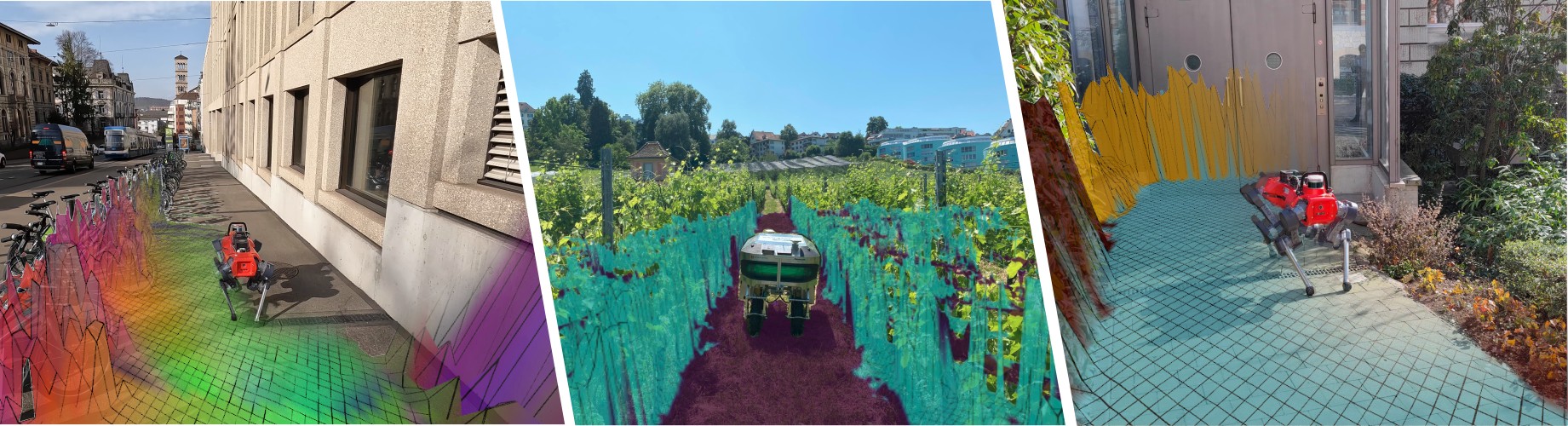}
                \caption{Three different applications of our multi-modal elevation mapping framework.
                Left: PCA layer of visual features on a crosswalk using LiDAR, Depth and RGB cameras.
                Center: semantic segmentation layer in an agricultural field using a RGB-D sensor.
                Right: semantic segmentation  layer in a garden using LiDAR, Depth and RGB cameras.
                }
        \end{minipage}
        \end{figure}
    \end{minipage}
    \label{fig:header}
\end{center}%
}]

\begin{comment}
\twocolumn[{%
\renewcommand\twocolumn[1][]{#1}%
\maketitle
\thispagestyle{empty}
\pagestyle{empty}
\begin{center}
    \centering
    %\begin{minipage}{\textwidth}
    %\begin{figure}[H]
    % left bottom right up
	\includegraphics[width=1.0\textwidth]{images/header_tmp.png}
	\caption{Semantic elevation mapping approach showcasing three different scenarios. The left and right images display a semantic segmentation layer of the  semantic elevation map in an agricultural and a forest setting respectively. The center image  showcases the semantic elevation map displaying a colorzed layer of the semantic elevation map.}
    %\end{figure}
    %\end{minipage}
    \label{fig:header}
\end{center}%
}]
\end{comment}
\begin{abstract}
Elevation maps are commonly used to represent the environment of mobile robots and are instrumental for locomotion and navigation tasks. However, pure geometric information is insufficient for many field applications that require appearance or semantic information, which limits their applicability to other platforms or domains. In this work, we extend a 2.5D robot-centric elevation mapping framework by fusing multi-modal information from multiple sources into a popular map representation. The framework allows inputting data contained in point clouds or images in a unified manner. To manage the different nature of the data, we also present a set of fusion algorithms that can be selected based on the information type and user requirements. Our system is designed to run on the GPU, making it real-time capable for various robotic and learning tasks. We demonstrate the capabilities of our framework by deploying it on multiple robots with varying sensor configurations and showcasing a range of applications that utilize multi-modal layers, including line detection, human detection, and colorization.
\end{abstract}
  \begingroup
  \renewcommand\thefootnote{}\footnote{$\dagger$~Authors are with the Department of Mechanical and Process Engineering, ETH Zürich, 8092 Zürich, Switzerland \{\href{mailto:gerni@ethz.ch}{\nolinkurl{gerni}}, \href{mailto:jonfrey@ethz.ch}{\nolinkurl{jonfrey}}, \href{mailto:tamiki@ethz.ch}{\nolinkurl{tamiki}}, \href{mailto:mahutter@ethz.ch}{\nolinkurl{mahutter}}\}\href{mailto:gerni@ethz.ch}{\nolinkurl{@ethz.ch}}. J. Frey is also associated with the Max Planck Institute for Intelligent Systems, 72076 Tübingen, Germany. $\ddagger$~M. Mattamala is with the Oxford Robotics Institute at the University of Oxford, UK \href{mailto:matias@robots.ox.ac.uk}{\nolinkurl{matias@robots.ox.ac.uk}}. This project was supported by the Max Planck ETH Center for Learning Systems (Frey), Swiss National Science Foundation (SNSF) through project 166232, 188596, the National Centre of Competence in Research Robotics (NCCR Robotics), and the European Union’s Horizon 2020 research and innovation program under grant agreement No.780883 and ANID / Scholarship Program / DOCTORADO BECAS CHILE / 2019-72200291 (Mattamala).}%
  \addtocounter{footnote}{-1}%
  \endgroup
  
\section{INTRODUCTION}
Autonomous unmanned ground vehicles deployed in outdoor environments need to understand their surroundings to operate in a safe and reliable manner. 
Onboard range sensors, such as LiDARs, stereo cameras, or RADARs can provide the robot with a spatial understanding of its surroundings. 
However, these different representations have varying nature and noise, which motivates their aggregation into a single unifying map.

Different types of maps have been developed for robotics. 
The simplest are 2D maps \cite{Thrun2006}, which store the occupancy information in each cell. 
3D volumetric maps are another popular representation~\cite{hornung13auro,oleynikova2017voxblox} that better encode the geometry of the environment, though at a higher computational and memory cost. 
2.5D maps (also called elevation or height maps)~\cite{Fankhauser2014, Fankhauser2018, Miki2022} are a compromise between both, and well suited for ground platforms.
All the aforementioned maps only contain geometric information about the environment. 
To successfully capture different environments, these representations need to be \emph{multi-modal}, here \emph{multi-modal} refers to different information content types i.e., encode semantic classes, friction, traversability, color, or other task-specific information, which is an active area of research nowadays~\cite{Ewen2022b, Gan2021a,grinvald2019volumetric,Rosinol20icra-Kimera}.
The multi-modal information can significantly enhance the performance of a variety of downstream tasks.
For example, semantic information enables robots to differentiate concrete from grass and mud, which allows them to plan a path on the less cost-intensive road surface.

% Contributions of the work
In this work, we aim to contribute to the development of multi-modal robotic perception by presenting a flexible, real-time capable \gls{mem} framework. 
It builds upon our previous work~\cite{Miki2022} to allow the seamless integration of multi-modal information such as geometry, semantics, and other data in various modalities from different sources. 
It is implemented on a \gls{gpu} to accelerate the heavy calculation of large data structures.
In addition, our framework provides customizable post-processing plugins as introduced in~\cite{Miki2022}, enabling the map to be adopted in a variety of tasks, including its use in learning approaches. 
This allowed us to develop different robotics and learning applications in a simple way, only using the single unified \gls{mem} framework.
Specifically, the contributions are:
\begin{itemize}
    \item A flexible multi-modal elevation mapping framework that can be fed with different representations such as point clouds and monocular images.
    \item Fusion strategies to deal with multi-modal data, such as geometry, RGB information, high-dimensional features, and semantic classes.
    \item Demonstration of multiple applications for robotics and learning that exploit the flexibility of the approach in terms of sensing and data modalities.
    \item Open source implementation of the system for the benefit of the community\footnote{\href{https://github.com/leggedrobotics/elevation\_mapping\_cupy}{https://github.com/leggedrobotics/elevation\_mapping\_cupy}}.
\end{itemize}

\begin{figure*}
  \includegraphics[width=1.0\textwidth]{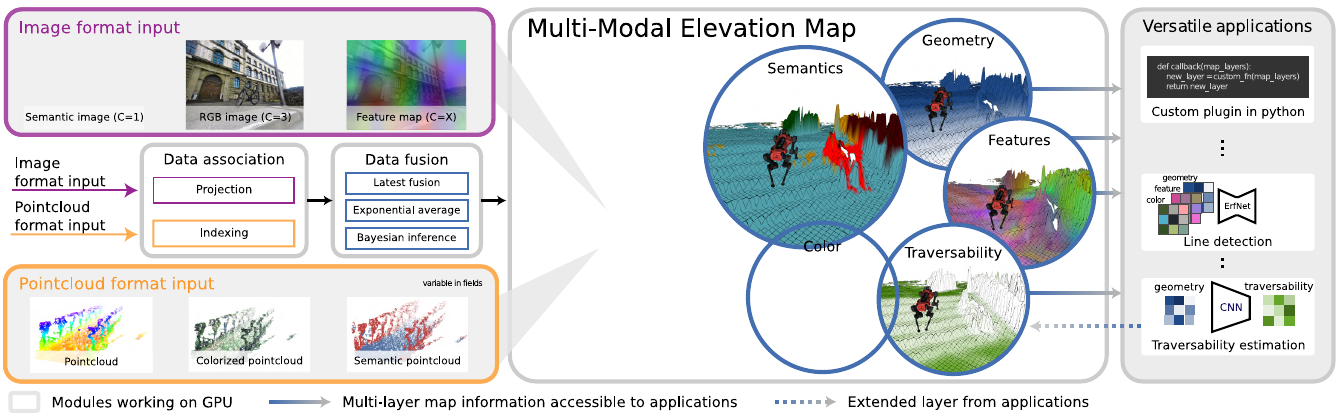}
      \caption{Overview of our multi-modal elevation map structure. 
      The framework takes multi-modal images (purple) and multi-modal (blue) point clouds as input.
      This data is input into the elevation map by first associating the data to the cells and then fused with different fusion algorithms into the various layers of the map.
      Finally the map can be post-processed with various custom plugins to generate new layers (e.g. traversability) or process layer for external components (e.g. line detection).}
    \label{fig:overview}
\end{figure*}

\section{RELATED WORK}

In this section, we review existing literature on map representations and semantic maps.
\subsection{Map representations}

First we review discrete representations for mapping, such as 2D occupancy maps, 2.5D elevation maps, and 3D volumetric representations.
2D occupancy maps are a common representation of the environment, that efficiently represent free and occupied space on a horizontal grid~\cite{Thrun2006}. 
They are suited for flat terrain and are computationally efficient.
However, they cannot capture uneven surfaces, making them unsuitable for complex terrains.
3D volumetric representations overcome this limitation and are able to represent the 3D complexity of the environment~\cite{Frey2022}, making them ideal for platforms such as drones, which can freely move in 3D space. 
Common approaches include octrees~\cite{hornung13auro} or voxel maps~\cite{oleynikova2017voxblox}. 
These maps are capable of handling overhanging obstacles, but creating volumetric maps comes at a cost of increased memory and computation.
For ground robots, such as wheeled and legged platforms, 2.5D representations such as elevation maps~\cite{Fankhauser2014} built from 3D sensing are a good compromise of expressiveness and efficiency, making them a popular choice for robotic applications.
These maps discretize the horizontal plane into cells, but store a continuous height value for each cell, allowing them to represent  complex and rough terrain such as staircases, steep slopes, or cluttered surfaces, which are particularly relevant for ground platforms. 
Multiple implementations of elevation mapping systems have been developed for \gls{cpu} systems, such as Fankhauser \etal~\cite{Fankhauser2018} and Ewen \etal~\cite{Ewen2022b}.
Pan et al.~\cite{Pan21} and Miki et al.~\cite{Miki2022} introduce a GPU-based implementation that enables real-time use of 2.5D elevation maps even with large amounts of data.
Our work builds upon the latter and extends it for RGB sensors, as well as adding new features to store semantic information.

\begin{table}[h!]
    \caption{\label{demo-table}Comparison of features of our work with prior work.}
    \begin{center}
        \begin{tabular}{c  c c c c c} 
            \toprule
            Feature & \gls{mem} & Elmap & Selmap  & Multitask  & Matur.  \\ [0.5ex] 
             &  &  \cite{Miki2022}&   \cite{Ewen2022b} &  \cite{Gan2021a} &  \cite{Maturana2018a} \\ [0.5ex] 
            % \hline
            \hline
            Class probability & \checkmark & \xmark & \checkmark & \checkmark & \checkmark \\ 
            % \hline
            Sem. Features & \checkmark & \xmark & \xmark & \checkmark & \xmark \\ 
            % \hline
            Real-time & \checkmark & \checkmark & \checkmark & (\checkmark) & \checkmark\\
            % \hline
            Monocular image & \checkmark & \xmark & \xmark & \xmark & \checkmark \\
            % \hline
            ROS & \checkmark & \checkmark & \checkmark & \checkmark & \checkmark \\
            % \hline
            GPU & \checkmark & \checkmark & \xmark & \xmark& ? \\ 
            % \hline
            Open source & \checkmark & \checkmark & \checkmark & \checkmark &\xmark\\ 
            \bottomrule
        \end{tabular}
    \end{center}
\end{table}

\subsection{Semantic Mapping}

To successfully perform certain tasks such as navigation and locomotion, purely geometric information does not suffice and semantic information is imperative \cite{Gan2021a}. 
Accumulation of the generated multi-task data in a map is desirable for downstream applications.
This has been done for 3D volumetric maps \cite{grinvald2019volumetric, Rosinol20icra-Kimera}, as well as for 2.5D semantic elevation maps \cite{Ewen2022b, Maturana2018a}.
In the following, we will review the semantic mapping approaches applied to ground robots and provide an overview of the existing methods and their features in \tabref{demo-table}.  

In \cite{Maturana2018a}, a robot-centric 2.5D elevation map is generated by fusing a LiDAR point cloud into the map.
A pixelwise semantic segmentation is extracted from images with a neural network and fused into the map using Bayesian inference. 
Hence, the map mainly contains fixed class probabilities given by the network. 
While our work is inspired by this fusion approach, we aim to extend the formulation to any kind of data modality.
Additionally, as shown in \tabref{demo-table}, this work is closed-source.

In \cite{Gan2021a} the authors present a multi-tasking neural network that takes RGB images as input and outputs class probabilities and traversability estimations.
The semantics in combination with their depth maps are fused into a  multi-layer semantic 3D volumetric map with a Bayesian inference approach~\cite{Gan20}, that leverages local correlations of the grid. 
This method runs on a \gls{cpu} and it is fixed to the map frame targeting large-scale semantic reconstructions.
This volumetric approach runs with logarithmic complexity.
While it is able to extend to new information modalities, new tasks need to be added to the multi-task network which requires retraining the entire custom network.
 As shown in \tabref{demo-table}, the method is not able to process monocular images containing multi-modal information.

In \cite{Ewen2022b} the authors present a method that generates a robot-centric elevation map that takes data from a single \mbox{RGB-D} camera as input and fuses the semantics into the map by associating the pixels to the elevation map with the depth information. 
To update the semantic information, a Bayesian inference technique is employed. 
This approach focuses on inputting class probabilities while we aim to provide a general framework that accepts multi-modal data and can potentially be extended for different platforms.

In this work, we extend our framework presented in ~\cite{Miki2022} to additionally capture multi-modal information.
We present a highly configurable GPU-based robot-centric elevation mapping, that can take multi-modal point clouds as input as well as additional data modalities from multi-modal images. The latter do not provide geometrical information but merely extend the modalities of the map. The \gls{mem} has already proven its effectiveness for fusing visual traversability in concurrent work~\cite{Frey2023}.
It runs in real-time and it is open source as shown in \tabref{demo-table}.

% from intro to check whther related work is complete
% An ideal mapping framework for various applications should be able to accept generic data types
% There is a dire need for real-time capability, especially with increased data modality.

\section{APPROACH}
The goal of this approach is to extend the elevation mapping presented in~\cite{Miki2022} by additionally fusing multi-modal information into the map.

\subsection{Overview}
\label{ssec:overview}

As shown in \figref{fig:overview}, the framework can be divided into three core modules: Data Association (\secref{ssec:data_assoc}), Fusion Algorithms~(\secref{subsec:fusion}), and Post-Processing~(\secref{ssec:post-processing}). 
Additionally, before inputting the sensor data into the core modules, the raw sensor data can be pre-processed to extract multi-modal information, such as visual features or semantic segmentations. 
The framework can take both multi-modal point clouds with additional fields as well as multi-modal images with custom channels as input.
These structures can contain multi-modal information such as class probabilities, RGB colors, and semantic features in each data point, which will be fused into the map.
In the Data Association module, the data is matched to the cells in the map either by associating each data point of the point cloud to a cell or by projecting each cell into the image. 
After establishing the correspondence between the data and the cells, the Fusion Algorithm updates the values of the cells by fusing the new information.
The map consists of pure geometric as well as multi-modal layers.
The latter are task-agnostic information containers and thus do not differentiate between different information modalities.
The  fusion algorithm is layer-specific and defines which information modality can be fused into these layers, as different modalities might need different fusion techniques. 
These two modules combined generate a unified manner to input multi-modal data into the map.
As a final step, the generated layers can be used for post-processing. 
The post-processing module consists of a plugin system where the user can choose which map layers to input and whether to generate new layers, modify the information within existing layers, or perform additional processing for external components.

\subsection{Data association}
\label{ssec:data_assoc}

\begin{figure}[h]
    \centering
    \includegraphics[width=\columnwidth]{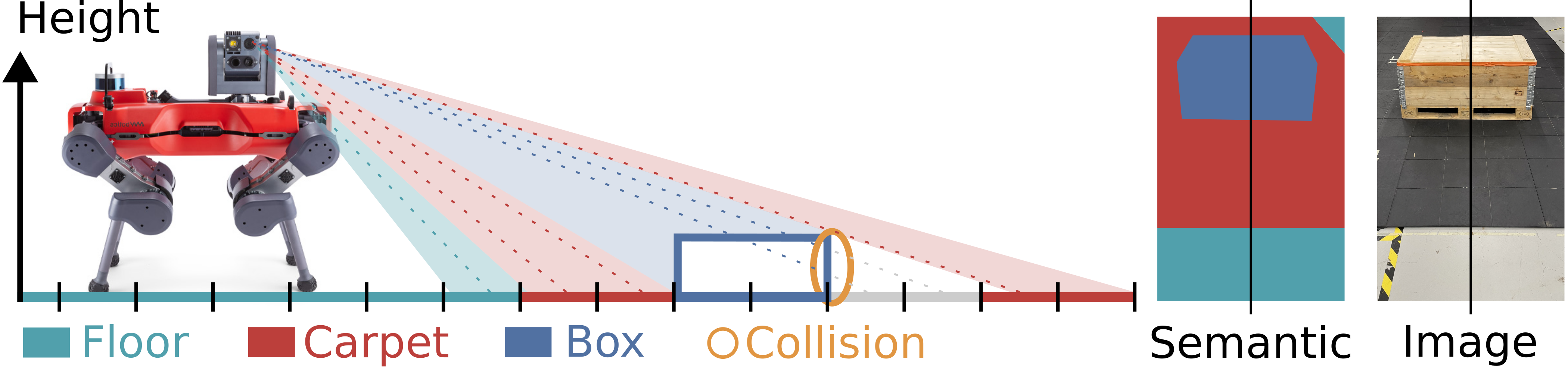}
    \caption{Association of pixel-wise semantic information to individual cells within the map. Each cell within the frustum of the corresponding camera is projected onto the image plane. An efficient ray-casting approach checks if the cell is visible or occluded by another cell based on the cell's height.}
    \label{fig:assoc}
\end{figure}
The proposed framework establishes correspondences between map cells and the data of a multi-modal point cloud or a monocular multi-modal image.
Multi-modal point clouds can be directly associated to cells according to the horizontal coordinates of each point in parallel.
The point cloud can be additionally used to update the elevation information according to \cite{Miki2022}.

In contrast, multi-modal images do not provide elevation information and we, therefore, rely on the elevation information accumulated within the map and a ray-casting schema for the data association shown in \figref{fig:assoc}. 
For most applications, the number of potentially visible cells within the robot-centric elevation map is smaller than the number of pixels contained within the image, therefore we choose to project the cells onto the image.
For each cell in the \gls{fov} of the camera, we evaluate if the cell is visible by checking the height of each intermediate cell along the ray connecting the camera's focal point and the cell. 
The intermediate cells are computed  using Bresenham's algorithm~\cite{bresenham}. 
A correspondence is successfully established between the cell and the image plane if, for all intermediate cells along the ray, the map elevation is smaller than the height of the ray connecting the camera focal point and the cell. 
In other words, the cell can be seen from the camera's perspective and is not occluded. 
The respective pixel coordinate in the image plane for the cell is computed using the pin-hole camera model with the known camera intrinsics and extrinsics.
This process is performed for each cell in parallel.

\subsection{Fusion algorithms}
\label{subsec:fusion}
Once  the correspondence of the new data and the  cell is defined, the semantic data is fused into the respective layers, updating their values.
To support different types of information and also a variety of use cases we implemented four different fusion algorithms: \emph{Latest} (\secref{sssec:latest}), \emph{Exponential Averaging} (\secref{sssec:exp_aver}), \emph{Gaussian Bayesian Inference} (\secref{sssec:bay_feat}), and \emph{Dirichlet Bayesian Inference} (\secref{sssec:dirichlet}).
The fusion algorithms can be configured by the user for each input source separately.
In addition, the user can select which layer of the \gls{mem} the data for each input source is fused to, allowing to fuse data from different sources into the same layer or in individual layers. 
The fusion algorithm needs to be selected based on the information modality since certain fusion algorithms cannot be applied to every modality.
The algorithms are applied to each layer of each data point in parallel. 
Given the modular design, new fusion algorithms can be added easily according to the user's needs.
Class probabilities can be input into the semantic mapping framework by either selecting a subset of all the classes or by inputting the top k classes.
The latter is convenient for memory optimization if there are many different classes.

\subsubsection{Notation}
\label{sssec:notation}
Vectors are written in bold and lowercase letters, and the index $i$ is used for measurement data points such as point cloud points and pixels. 
The subscript $j$ is used to index elevation map cell elements and $t$ is the time step of an update.
Each point cloud consists of $N$ data points and for cell $c_j$ there are $N_j$ data points that correspond to that cell.
The class information can be represented with class likelihood $\mathbf{p}\in \mathbb{R}^K$ where $p_k \in [0,1]$ where $k$ is the class index and $\sum_k^K p_k =1$ and $K$ is the total amount of classes, or in one-hot encoded vectors $\mathbf{p}\in \mathbb{R}^K$ where $p_k \in \{0,1\}$  and $\sum_k^K p_k=1$.
$\theta_{j}$ is the value of a layer of cell $j$.
Features are vectors associated with the image content at a certain location of a pixel or of a 3D point.
The extracted features are encoded in $d$-dimensional vectors $\mathbf{f} \in \mathbb{R}^d$.

\subsubsection{Latest}
\label{sssec:latest}
A straightforward method to fuse the information is to only retain the latest measurement per cell.
For the case of point clouds, it might be that multiple data points fall into one cell.
In that case the information is fused into the semantic elevation map by layer-wise averaging all the data points that fall within the same cell $c_j$ according to:
\begin{equation}
    \mathbf{a_{t,j}}= \frac{1}{N_j}\sum_{i=0}^{N} \mathbf{m_i}\cdot  \mathds{1}_ {\{\mathbf{m_i} \in c_j\}}
    \label{eq:average}
\end{equation}
where $\mathbf{m_i} $ is the $i^{th}$ observation.

This fusion method can be used for all different information modalities.
However, it does not consider any past measurements making it more prone to noise.

\subsubsection{Exponential averaging}
\label{sssec:exp_aver}
Exponential averaging consists of assigning greater weight to more recent data and exponentially decreasing the significance of previous data as shown in \figref{fig:fus_alg}b.
This method improves the robustness towards noise by considering previous measurements.
For point clouds, the algorithm first computes the average of all the measurements that correspond to a cell $c_j$ according to \eqref{eq:average}. Then the new cell value is computed as:
\begin{equation}
    \boldsymbol{\theta_{t,j}}=w \cdot \mathbf{a_{t,j}} + (1-w) \cdot \boldsymbol{\theta_{t-1,j}}
\end{equation}
where $w$ is a user-defined weight.
This fusion technique can be used for all semantic information types as it can be accommodated to work with real numbers $\mathbb{R}$ as well as probabilities $\mathbb{R}\in [0,1]$. 
Since exponential averaging does not follow a probabilistic approach, the information stored in the layer cannot be regarded as a probability measure.
This motivates the two following probabilistic fusion algorithms.

\subsubsection{Bayesian inference of Gaussian distributions}
\label{sssec:bay_feat}
This method is suited for $d$-dimensional data $\boldsymbol{f} \in \mathbb{R}^d$ and uses Bayesian inference with the assumption that the data 
$\boldsymbol{f_i}$ is drawn from a Gaussian distribution with known variance. 
The likelihood function is the probability of the observations given the mean.
\begin{equation}
    p(\mathcal{F}|\boldsymbol{\mu_{f,j}})=\mathcal{N}(\mathcal{F}|\boldsymbol{\mu_{f,j}},\boldsymbol{\sigma^2_{f,j}}\mathit{I})
\end{equation}
where $\boldsymbol{\mu_{f,j}},\boldsymbol{\sigma^2_{f,j}}$  are the parameter vectors describing the cell's mean and variance and  $\mathcal{F}=\{\boldsymbol{f_0},...,\boldsymbol{f_{N_j}}\}$ is the set of $N_j$ observations.
We choose the prior to be Gaussian distributed:
  \begin{equation}
      p(\boldsymbol{\mu_{f,j}})=\mathcal{N}(\boldsymbol{\mu_{f,j}}|\boldsymbol{\mu_{0,j}},\boldsymbol{\sigma^2_{0,j}}\mathbf{I})
  \end{equation}
  where $\boldsymbol{\mu_{0,j}},\boldsymbol{\sigma^2_{0,j}}$ are the parameter vectors describing the prior's mean and variance for cell $j$.
This prior is a conjugate prior with respect to the likelihood, thus the posterior will also be a Gaussian distribution, and allows for closed form solution for the posterior using Bayes rule \cite{bishop2006}:
\begin{equation}
    p(\boldsymbol{\mu_{f,j}}|\mathcal{F}) = \mathcal{N}(\boldsymbol{\mu_{f,j}}|\boldsymbol{\mu_{N,j}},\boldsymbol{\sigma^2_{N,j}}\mathbf{I})
\end{equation}
where 
\begin{equation}
    \label{eq:men_n}
    \mu_{N,d} = \frac{\sigma_{f,d}^2}{N\sigma_{0,d}^2+\sigma_{f,d}^2}\mu_{0,d} + \frac{N\sigma_{0,d}^2}{N\sigma_{0,d}^2+\sigma_{f,d}^2}\mu_{ML,d}
\end{equation}
\begin{equation}
    \label{eq:sig_n}
    \sigma_{N,d} = \frac{\sigma_{f,d}^2\sigma_{0,d}^2}{N\sigma_{0,d}^2+\sigma_{f,d}^2}
\end{equation}
where $\mu_{ML,d}=\frac{1}{N_j}\sum _{i=1}^N f_{i,d}\cdot  \mathds{1}_ {\{\mathbf{f_i} \in c_j\}}$.\\
In equations \eqref{eq:men_n} and \eqref{eq:sig_n}, we neglect the cell's index $j $ for readability purposes and $d$ indicates the index of the dimension.
The closed-form solution allows for real-time updating of the cell values.
This method, as previously mentioned, is used for real-valued data $\mathbb{R}^d$ and thus it is suited for e.g. features.

\subsubsection{Bayesian inference of Dirichlet distributions}
\label{sssec:dirichlet}
This fusion algorithm is suited for class probabilities.
Cell-wise class probabilities are represented by a categorical distribution $\boldsymbol{\theta_{j}} \in  [0,1]^K $, where each element represents the probability of cell $j$ belonging to class $k$ from a fixed set of $K$ classes $\mathcal{K}=\{1,2,...,K\}$.

\begin{figure}[h]
    \centering
    \includegraphics[width=\columnwidth]{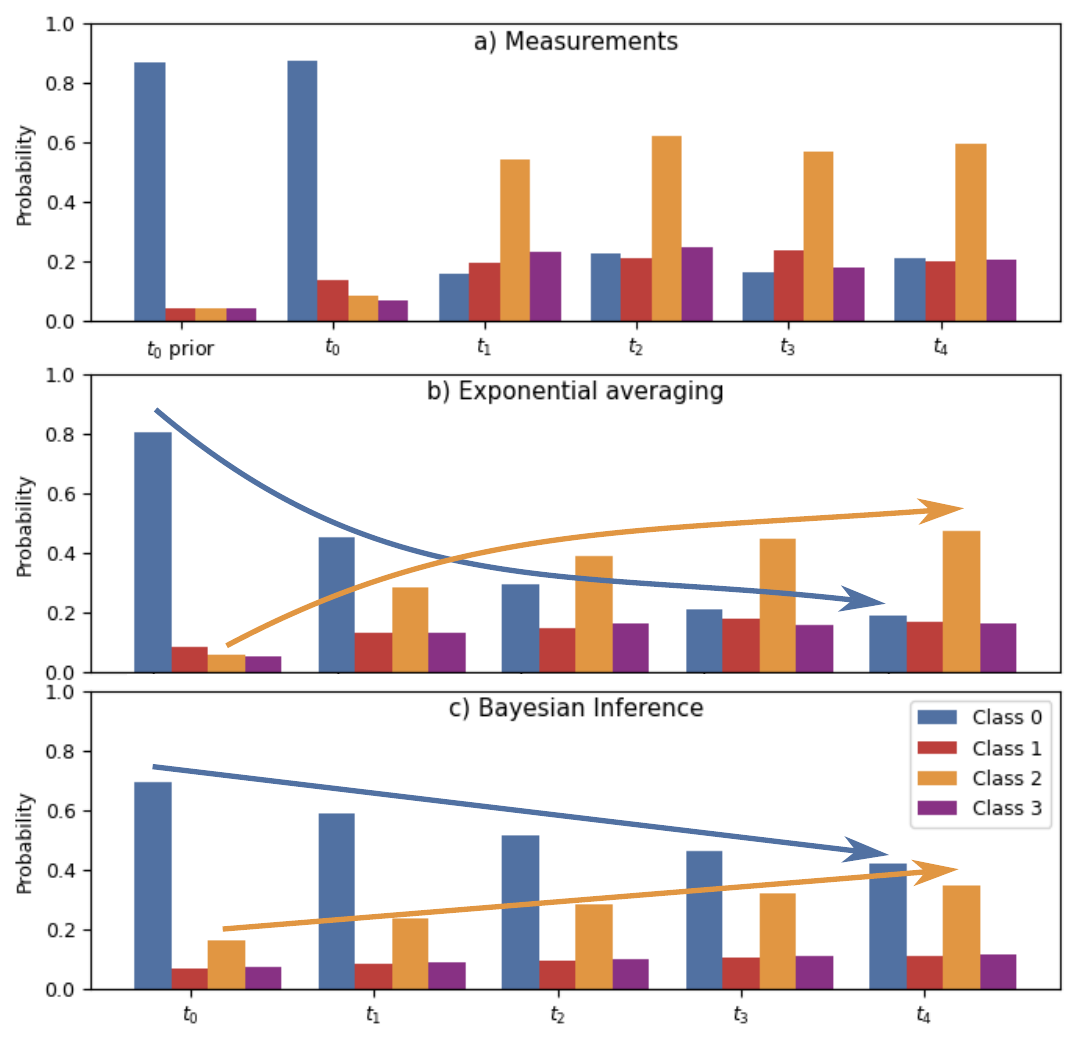}
    \caption{Fusion algorithms behaviour over time. In Figure (a) the prior at $t_0$ and the measurement that falls into one cell is shown. (b) depicts the exponential decay of Class 0 confidence as a result of exponential averaging fusion. (c) illustrates the Bayesian inference fusion method, which exhibits a more gradual decrease in the Class 0 confidence.}
    \label{fig:fus_alg}
\end{figure}

The likelihood of the map cell is described with a categorical distribution $p(\mathbf{m_i}|\boldsymbol{\theta_{j}})=\prod_{k=1}^K {\theta_{j,k}}^{m_{i,k}}$, where $\mathbf{m_i}=[m_{i,1},...,m_{i,K}]$ is the semantic measurement.
Thus the likelihood over a set of $\mathcal{D}$ observations is:
\begin{equation}
    p(\mathcal{D}|\boldsymbol{\theta_{j}})=\prod_{k=1}^K {\theta_{j,k}}^{m_k}
\end{equation}
where $m_k = \sum^{N_j}_i m_{i,k}$ and the value $N_j$ is the number of measurements that fall within the cell $j$.\\
Our goal is to seek the posterior over $\boldsymbol{\theta_{j}}$.
Similar to \secref{sssec:bay_feat} we choose a conjugate prior, in this case, the Dirichlet distribution:
\begin{equation}
    p(\boldsymbol{\theta_{j}}|\boldsymbol\alpha_{j})= \frac{\Gamma(\sum_{k=1}^K \alpha_{j,k})}{\prod_{k=1}^K \Gamma(\alpha_{j,k})}\prod_{k=1}^K \theta_{j,k}^{\alpha_{j,k}-1}\propto \prod_{k=1}^K \theta_{j,k}^{\alpha_{j,k}-1}
\end{equation}
where $\Gamma = \int_0^{\infty} t^{\alpha_{j,k}-1} e^{-t} dt $ and $\boldsymbol{\alpha_j}$ are the parameters of the distribution. 
The resulting posterior is a categorical distribution:
\begin{equation}
    p(\boldsymbol{\theta_{j}}|\mathcal{D}, \boldsymbol\alpha_{j})\propto p(\mathcal{D}|\boldsymbol{\theta_{j}}) \cdot p(\boldsymbol{\theta_{j}}|\boldsymbol\alpha_{j}) \propto \prod_{k=1}^K \theta_{j,k}^{\alpha_{j,k}+m_k-1},
\end{equation}
with the closed-form solution being:
\begin{equation}
    \label{eq:posterior}
    p(\boldsymbol\theta_j|\mathcal{D},\boldsymbol\alpha_j)=\frac{1}{\sum_{k=1}^K \alpha_{j,k}} \boldsymbol\alpha_j
\end{equation}
\begin{equation}
    \boldsymbol\alpha_{j,t}= \boldsymbol\alpha_{j,t-1}+\sum^{N_j}_i \mathbf{m_{i}},
\end{equation}
following the derivation by Bishop~et al.~\cite{bishop2006}.
The posterior is the updated class probability for each cell. 
This Bayesian inference approach can be used for $K$ different classes and keeps track of all past observations as shown in \figref{fig:fus_alg}.

\subsection{Post processing}
\label{ssec:post-processing}
\begin{figure*}[h]
    \centering
    \includegraphics[width=1.0\textwidth]{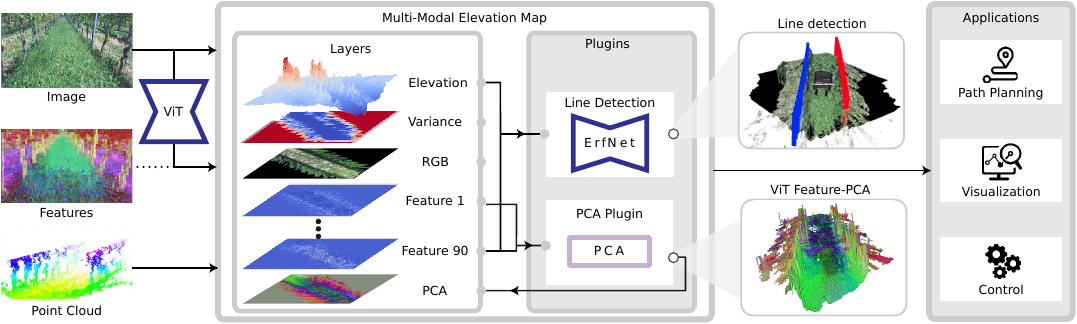}
    \caption{Post-processing. This figure shows how the map can be employed to generate insight and a new layer. 
    On the left, the input image and the feature extraction step are displayed.
    This data is fed into the elevation map in the middle of the figure. 
    The top right side displays a post-processing plugin that generates the the left (red) and right (blue) tree lines predictions of a vineyard.
    On the lower right side a plugin takes feature layers as input and generates a PCA layer. }
    \label{fig:post_proc}
\end{figure*}
 
Once all the sensor information is contained in the map, the framework allows to post-process the map layers to generate task-relevant data.
These updates are triggered by an event and thus run at a user-defined frequency.
The different post-processing options are implemented as plugins that can be configured by the user with a few lines of code, as described in \cite{Miki2022}.

We enhance this system by also providing additional layers such as RGB or semantics to the plugins.
These plugins can utilize the layer information that is already on \gls{gpu}, thus reducing the data transferring time. 
We demonstrate the usability with several examples such as neural networks that take map layers as input and generate a new layer (\figref{fig:post_proc}b), modify existing layers within the \gls{mem}, or perform additional processing for external components (\figref{fig:post_proc}a).
Please refer to \secref{sec:experiments} for a detailed description.

\section{EXPERIMENTS}
\label{sec:experiments}
We first show the implementation details of the framework (\secref{ssec:implementation}).
We then present the results by profiling the time performance of the implementation (\secref{ssec:perf_analysis}). 
Finally, we showcase how our \gls{mem} framework can benefit a variety of robotic applications with multiple case studies performed on different robots 
(\secref{ssec:color_app}, \secref{ssec:sem_seg_app}, \secref{sec:lineparam}).

\subsection{Implementation}
\label{ssec:implementation}
The map is implemented in Python, using CuPy \cite{Okuta2017CuPyA} which allows \gls{gpu} acceleration and the generation of user-defined low-level \gls{cuda} kernels for customized highly parallel data processing. 
It uses ROS \cite{ros} for inter-process communication and the core modules implemented in Python are wrapped by a C++ framework to accelerate the serialization of ROS data.
The map is published at a user-defined rate as a GridMap message \cite{Fankhauser2016GridMapLibrary}.

\subsection{Performance analysis}
\label{ssec:perf_analysis}

For the performance analysis, we run the elevation mapping framework with a $\SI{10}{m} \times \SI{10}{m}$ grid with a resolution of $\SI{4}{cm}$ resulting in a grid of $250\times250$ cells. 
The map is tested on a robot with one Stereolabs ZED 2i camera, publishing an image of $360 \times 640$ pixels at a framerate of \SI{3}{Hz}.
The image stream is processed with a pre-trained semantic segmentation network~\cite{lraspp}.
All the time profiling tests were performed over 300 iterations.

\renewcommand{\arraystretch}{1.3} 

\begin{table}[h!]
    \caption{Table containing the time performance of the semantic elevation map processing a point cloud of 230'400 points. }
    \centering
    \begin{tabular}{ l c c  }
        \toprule
        Description & RTX 4090 [\SI{}{ms}] &  Jetson Orin [\SI{}{ms}]   \\ 
        \hline
        point trsf. \& z err. cnt & $0.047 \pm 0.013 $  &  $0.695\pm0.246$ \\  
        drift compensation & $0.108 \pm 0.093$ & $ 0.915\pm 0.551$  \\  
        height update \& ray cast & $1.341 \pm 0.107$ & $ 17.524\pm 2.417 $  \\  
       \emph{multi-modal update}  & $0.306 \pm 0.115$ &  $1.536\pm 0.508$  \\  
        overlap clearance & $0.226 \pm 0.100$ &  $0.970\pm 0.186$  \\  
        traversability & $0.538 \pm 0.159$ &  $1.936\pm 0.200$ \\  
        normal calculation & $0.030 \pm 0.030$ &  $0.060\pm 0.112$  \\  
                \hline

        \textbf{Total} &  $2.596\pm0.260$  &  $23.635\pm 2.754$   \\
        \bottomrule
    \end{tabular}
    \label{table:perf}
\end{table}

We measure the time performance of the \gls{mem} update as shown in \tabref{table:perf}, following \cite{Miki2022}. 
The measurements are performed on a workstation with a GeForce RTX 4090 \gls{gpu} and Nvidia Jetson Orin.
The total time required for the entire update is \SI{2.6}{ms} / \SI{23.6}{ms} corresponding to an update rate of \SI{385.2}{Hz} / \SI{42.3}{Hz} on the respective hardware.
Fusing the points into the elevation map (height update \& ray cast) is the step that takes the most time as it needs to fuse a high number of points into the map.

Comparing the time performance with \cite{Ewen2022b}, we can notice that even when having four times more cells, we run at approximately the same speed.
By adding multi-modal data to the elevation map \cite{Miki2022}, the time performance of the map stays in the same order of magnitude, making it real-time capable.
\begin{table}[h!]
    \caption{Point cloud sizes of different sensors. }
    \centering
    \begin{tabular}{ l c c  }
        \toprule
        Sensor & Size &  Frequency   \\ 
        \hline
        ZED 2i & 230'400  & \SI{3}{Hz} \\  
        Velodyne VLP-16  & 300'000  & \SI{1}{Hz} \\
        Depth cameras &  407'040 &  \SI{15}{Hz} \\
        \bottomrule
    \end{tabular}
    \label{table:pcl}
\end{table}

\begin{figure}[h]
    \centering
    \includegraphics[width=0.48\textwidth]{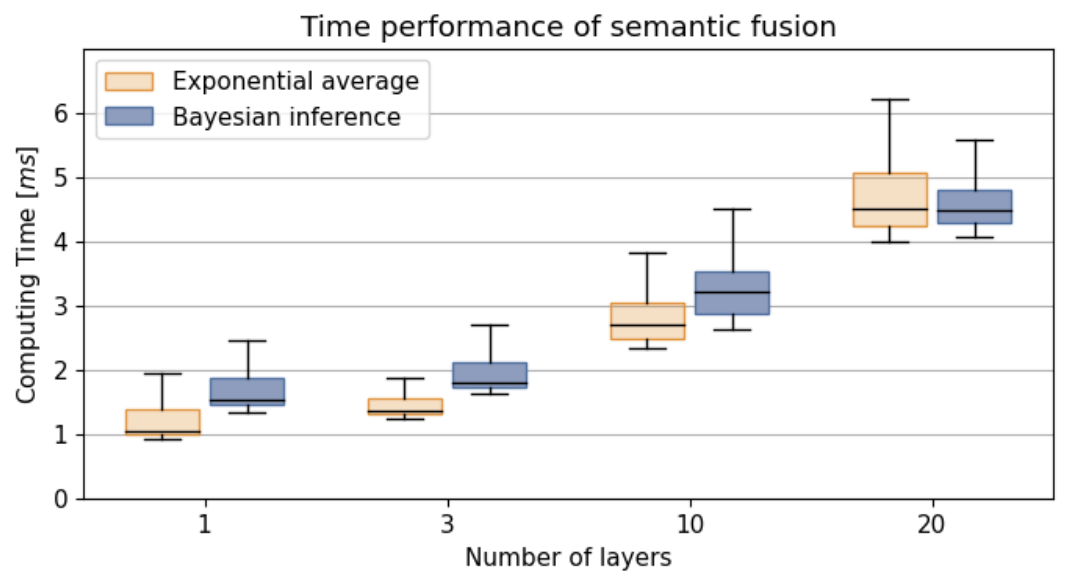}
    \caption{Time performance of the semantic fusion step with increasing numbers of layers recorded on a Jetson Orin.}
    \label{fig:time_perf}
\end{figure}

We test the scalability of the approach by running the framework on a Jetson Orin with the previously presented configuration with an increasing number of multi-modal layers.
In \figref{fig:time_perf}  we show that the computational time of multi-modal update increases linearly to the number of multi-modal layers.  
The total processing time of the Jetson Orin for 20 layers is \SI{28.9}{ms} for exponential averaging and \SI{29.0}{ms} for Bayesian inference.

Finally, we inspect the time performance of the point cloud creation, which consists of computing the 3D coordinates of the points starting from the RGB-D setup presented in \secref{ssec:perf_analysis} which publishes images at a rate of \SI{3}{Hz}, processing the images, and publishing the point cloud. 
The bottleneck in this configuration proves to be running the image through a Lite R-ASPP Network with a MobileNetV3-Large backbone ~\cite{lraspp} that takes $\SI{31}{ms}$.
However, given that our framework does not depend on the specific network the user can select the network according to the time and performance requirements.

Our approach is memory efficient.
The point cloud with $100'000$ points with x, y, z, RGB, and 5 semantic channels solely needs \SI{7.2}{MB} of memory.
The 2.5D \gls{mem} with $200\times200$ cells and the same amount of semantic channels requires \SI{3.8}{MB} of memory.
Hence, we only require  \SI{1.6}{MB} of additional memory space in comparison to the elevation map of \cite{Miki2022} which allocates \SI{2.2}{MB}.
Additionally, as described in \ref{sssec:dirichlet}, the memory consumption can be further reduced by choosing the solution where we capture the top k class probabilities.

\subsection{Colorization from multi-modal data}
\label{ssec:color_app}

\begin{figure}[h]
\begin{minipage}[H]{0.48\textwidth}
    \includegraphics[width = \textwidth]{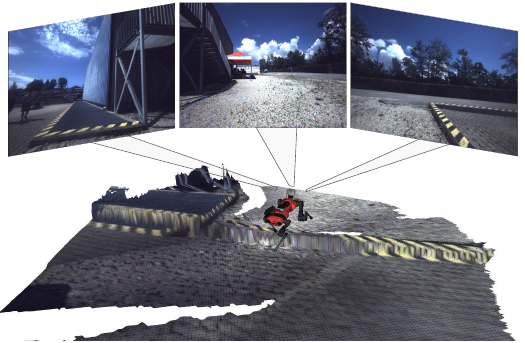}
\end{minipage}
\hfill
\begin{minipage}[H]{0.24\textwidth}
\end{minipage}
\caption{Color layer: The semantic elevation map is used in an outdoor environment with three point clouds and three images as input and displays the color layer.}
\label{pics:color}
\end{figure}
Our framework is able to fuse color information from different data modalities, whereas previous approaches always required a colorized point cloud or a depth-aligned RGB image.
This is particularly useful for robots with a diverse set of sensors.
In \figref{pics:color}, a qualitative analysis shows that the map can accurately represent the colors of the environment.
The sensing setup consists of a Sevensense Alphasense Core camera unit with three monocular color cameras with global shutter, that capture images at a frame rate of \SI{9}{Hz} with a resolution of $1080\times1440$ pixels and four Intel RealSense D435 depth cameras available on the ANYbotics ANYmal C legged robot, capturing images at a frame rate of \SI{15}{Hz} with a resolution of $480\times848$ pixels.
The layer can be used to improve the interpretability of the map by humans, as well as for downstream learning tasks.

\subsection{2.5D Semantic Segmentation}
\label{ssec:sem_seg_app}

\begin{figure}[h!]
\centering
    \includegraphics[width=\columnwidth]{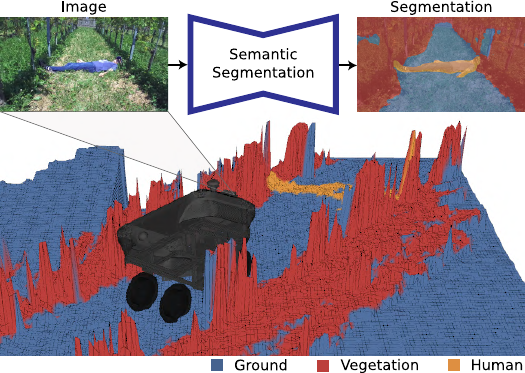}
\caption{Semantic segmentation mask layer of the elevation map. The colors of the elevation map encode the highest class probabilities. The human lying in the high grass is not distinguishable using the height (geometric) information but the semantics clearly detect the person (orange).}
\label{pics:semseg}
\end{figure}

In this second application, we demonstrate that the framework can be used to project class probabilities into the map.
In this specific case study, we use a similar setup as described in \secref{ssec:perf_analysis} where a stereo camera with a resolution of $360 \times 640$ pixels is used. 
The image is processed with Detectron2 \cite{Detectron2018} which outputs pixel-wise class probabilities. 
The class probabilities and the depth map are used to generate a semantic point cloud.
These points are then fused into the \gls{mem} according to the Bayesian inference of Dirichlet distributions described in  \secref{sssec:dirichlet}.
We showcase the pipeline in an agricultural setting as shown in \figref{pics:semseg}.
The figure depicts a person lying in the grass, who is identified by the segmentation network.
A purely geometric perception of the environment is not sufficient to recognize the human in the high grass. 
By integrating the semantics into the elevation map it can be used by a local planning module, which can take into account geometric as well as semantic obstacles.
This example showcases the importance of multi-modal and more specifically semantic information for real-world robotic applications.

\subsection{Line detection in agricultural setting}
\label{sec:lineparam}
We show that the proposed framework can be extended to take features as input, fuse them into the map, and train a custom network that takes multiple layers as input. 
Specifically, we showcase visual features representing latent embeddings of RGB images in an agricultural setting where we predict the tree lines in special crops as shown in \figref{fig:post_proc}a.
The described framework is tested on a wheeled robot with one RGB-D camera, publishing an image of $360\times640$ pixels.
In this application, we extract features from the RGB image with a self-supervised pre-trained vision transformer (ViT) \cite{CaronTMJMBJ21}.
We fuse these features according to the Bayesian inference of Gaussian distributions described in \secref{sssec:bay_feat}.
In this use case, we use elevation and semantic information as input to an end-to-end learnable network that predicts two tree lines in special crop fields.
The goal of the model is to predict the line parameters in the robot's frame. 
The lines are modeled as second-order polynomials.
The machine learning pipeline consists of two sequential blocks and follows the work of \cite{linedet}.
The first block is a \gls{cnn} that generates weight maps from the elevation and feature data. 
The second block takes the weight maps as input and predicts the line parameters of the tree line.
The \gls{cnn} consists of the  \gls{erf}. 
It predicts a weight map with 2 channels, one for each predicted line. 
The output of the \gls{erf} is fed into  the second block consisting of a least squares layer (LSQ) predicting the line parameters.
The predicted lines resulting from the approach can enable robot navigation, even in the presence of state estimation inaccuracies.

\section{CONCLUSION}
In this work we extended a state-of-the-art 2.5D elevation map by including multi-modal information of the environment, in order to provide an easy-to-use and computationally efficient tool for various robotics and learning tasks.
We presented a highly configurable framework that allows the user to configure the map content and enables a great variety of downstream tasks. 
We have shown high performance and low memory consumption of our \gls{mem} framework running on \gls{gpu}. 
To demonstrate that our approach can be used for various use cases, we have shown three different example applications.
% case studies.
We showed that the RGB colors can be injected into the map, and we demonstrated that various pre-trained semantic segmentation networks can be used to update the layers.
Finally, we introduced a learning-based tree line detection of special crops which uses the feature and elevation layers of the \gls{mem} as input.
We hereby demonstrate that our \gls{mem} framework serves as a highly configurable environmental representation tool which is easy to use and openly accessible for the research community.

%\bibliography{references}
{\small
\balance
\bibliographystyle{IEEEtranS}
\bibliography{bibliography/automatic,bibliography/manual}
% \bibliography{bibliography/manual}
}

\end{document}